\documentclass[letterpaper]{article} 
\usepackage{aaai2026}
\usepackage{times}
\usepackage{amsmath}   
\usepackage{amssymb}   
\usepackage{amsfonts}
\usepackage{amsthm}     
\usepackage{helvet}  
\usepackage{courier}  
\usepackage[hyphens]{url}  
\usepackage{graphicx} 
\usepackage{natbib}  
\usepackage{caption} 
\usepackage{algorithm}
\usepackage{algorithmic}
\usepackage{newfloat}
\usepackage{listings}
\usepackage{xspace}
\usepackage{adjustbox}

\usepackage{booktabs}
\usepackage{multirow}
\usepackage{subcaption}
\usepackage{booktabs}
\usepackage{helvet}
\usepackage{amsmath}  
\usepackage{subcaption} 
\usepackage{mathtools} 
\usepackage{bm}             
\usepackage{enumitem} 
\makeatletter
\renewcommand{\copyright@text}{}
\makeatother

\DeclareCaptionStyle{ruled}{labelfont=normalfont,labelsep=colon,strut=off} 

\newtheoremstyle{bolddef}
  {3pt}{3pt}          
  {}                  
  {}                  
  {\bfseries}         
  {.}                 
  { }                 
  {}                  

\theoremstyle{bolddef}
\newtheorem{definition}{Definition}

\theoremstyle{plain} 
\newtheorem{lemma}{Lemma}
\newtheorem{proposition}{Proposition}
\newtheorem{theorem}{Theorem}

\floatstyle{ruled}
\newfloat{listing}{tb}{lst}{}
\floatname{listing}{Listing}
\title{Unsupervised Learning for the Elementary Shortest Path Problem}
\author{
   Jingyi Chen\textsuperscript{\rm 1},
   Xinyuan Zhang\textsuperscript{\rm 1},
   Xinwu Qian\textsuperscript{\rm 1}\thanks{Corresponding Author}\\
}
\affiliations{
    \textsuperscript{\rm 1}Department of Civil and Environmental Engineering, Rice University\\

    jc358@rice.edu, xz125@rice.edu, xq15@rice.edu\\
}
\usepackage{bibentry}

\begin{document}

\maketitle

\begin{abstract}
The \emph{Elementary Shortest‑Path Problem} (ESPP) seeks a minimum cost path
from $s$ to $t$ that visits each vertex at most once. The presence of negative-cost cycles renders the problem $\mathcal{NP}$‑hard. We present a probabilistic method for finding near-optimal ESPP, enabled by an unsupervised graph neural network that jointly learns node value estimates and edge-selection probabilities via a surrogate loss function.
The loss provides a high probability certificate of finding near-optimal ESPP solutions by simultaneously reducing negative-cost cycles and embedding the desired algorithmic alignment. At inference time, a decoding algorithm transforms the learned edge probabilities into an elementary path. Experiments on graphs of up to 100 nodes show that the proposed method surpasses both unsupervised baselines and classical heuristics, while exhibiting high performance in cross-size and cross-topology generalization on unseen synthetic graphs.
\end{abstract}


\section{Introduction}
Given a weighted directed graph \( G= (V, E, w)\), the \textit{Elementary Shortest-Path Problem} (ESPP) seeks a minimum‑cost $s-t$ path from a source $s$ to a sink $t$, visiting each vertex at most once. It reduces to the standard shortest path problems (SPP) if the graph is acyclic. However, in general graphs when negative-cost cycles (NCCs) are present, solving ESPP is known to be $\mathcal{NP}$-hard~\cite{taccari2016integer}. The elementary visitation constraint invalidates the traditional dynamic programming (DP)-based algorithm for SPP~\cite{drexl2014solving}. As a result, exact solutions for ESPP rely on labeling algorithm whose state spaces grow exponentially in $|V|$~\cite{feillet2004exact} and shares the same worst-case computational complexity as the Traveling Salesman Problem (TSP) with profits \cite{feillet2005traveling}. 

While challenging, ESPP carries significant practical and theoretical implications. Practically, it arises naturally in transportation systems where a subset of edges offers one-time subsidies, which imposes elementary condition and introduces negative cycles. More fundamentally, ESPP forms the cornerstone of branch-and-price algorithms widely used for solving large-scale routing and scheduling problems~\cite{chabrier2006vehicle,amaldi2014maximum}. In these methods, ESPP instances are repeatedly solved as pricing subproblems, and optimality is established when ESPP solutions are non-negative. Consequently, unlike many other routing problems for which heuristic solutions may suffice, \textit{ESPP explicitly requires efficient algorithms with provable performance guarantees}.

Recent advances in machine learning, especially Graph Neural Networks (GNNs), have driven interest in solvers for hard combinatorial optimization problems \cite{smith1999neural,kool2018attention,bengio2021machine,gasse2022machine,khalil2022mip,cappart2023combinatorial}, inspiring a new wave of learning heuristics. Previous successes in solving combinatorial optimization problems such as the TSP have primarily relied on supervised and reinforcement learning approaches \cite{fu2021generalize,hudson2021graph,zheng2021combining,joshi2022learning,cheng2023select,pan2023h}. Supervised pipelines fit neural models to optimal or near‑optimal solutions, but generating such labels quickly becomes intractable as graph size grows. Reinforcement learning typically relies on non‑differentiable policy updates, resulting in high computational cost and training instability. Both reinforcement learning and supervised learning can lead to poor generalization on large graphs~\cite{hu2024assessing,yang2025aria}.

Unsupervised learning aids in generalization and offers a promising escape from the label bottleneck. In practice, however, designing a differentiable objective whose minima align with both feasible and near-optimal solutions is challenging. Recent approaches have shown the promise of unsupervised learning GNN to guide neural models over some combinatorial optimization problems \cite{karalias2020erdos,min2023unsupervised}. In particular, they adopt the probabilistic method \cite{karalias2020erdos}, which constructs a learned distribution that maximizes the probability of sampling a low-cost and feasible solution. This approach offers valid solutions with polynomial-time sampling complexity. However, these methods focus on modeling the combinatorial structure directly, often introducing bilinear interactions between node variables. For ESPP, however, directly modeling the problem will involve an exponential number of flow-based constraints such as subtour elimination constraints \cite{taccari2016integer}. Relaxing those elementary constraints and incorporating them into surrogate loss functions remains an open challenge in neural optimization.  

In light of the above observations and limitations, we propose a new learning framework that sidesteps the need to explicitly encode exponentially complex combinatorial structures. First, instead of introducing bilinear interactions between node variables, we independently learn edge-level probabilities and node value estimates, which improves stability and convergence during the training process. Second, we avoid exponential subtour elimination constraints by learning to construct subgraphs without negative cycles, enabling the use of efficient shortest path algorithms. Finally, inspired by the principle of algorithmic alignment \cite{xu2019can}, we embed algorithmic inductive bias in both the architecture and loss, guiding the model toward valid edge sequences and improving generalization in ESPP.


This study is an initial work by introducing \textsc{Espp-nnaa}, an unsupervised graph‑neural-network solver for the ESPP. Here, \textsc{Espp-nnaa} stands for \textbf{E}lementary \textbf{S}hortest \textbf{P}ath \textbf{P}roblem via \textbf{N}on-\textbf{N}egative cycle with \textbf{A}lgorithm \textbf{A}lignment. We summarize our contributions as follows:
\begin{itemize}
    \item We propose a novel probabilistic method to solve ESPP via unsupervised learning, bypassing exponentially many constraints by constructing a tractable subspace aligned with the standard shortest path problem.
    \item  We provide high probability certificates of finding near-optimal ESPP using the proposed probabilistic method. 
    \item We conduct extensive computational experiments demonstrating that \textsc{Espp-nnaa} achieves superior performance over state-of-the-art baselines and effectively generalizes from small training graphs to larger graphs with different topological structures.
    
\end{itemize}
\section{Preliminaries}
\subsubsection{Probabilistic Method}
An effective solution to mitigate the computational challenges of ESPP is to approach the problem from the lens of the probabilistic method (PM)~\cite{alon2016probabilistic}. Specifically, if an oracle support distribution $D$ over the edge set $E$ is available, one can sample low-cost and feasible solutions from $D$ in polynomial time.
The difficulties, however, arise from constructing such an oracle. When the underlying mathematical problem with objective function $f$ and constraints $\Omega$ can be analytically expressed,
one can train \(g_\theta\) to yield a solution distribution \(\mathcal{D}\) where a surrogate loss certifies that the distribution contains valid solutions.~\cite{karalias2020erdos}.
Specifically, we construct such surrogate loss function to maintain feasibility via a penalty for constrained problems:
\begin{equation*}
\ell(\mathcal{D}; G) = \mathbb{E}_{S \sim \mathcal{D}}[f(S; G)] + \beta \cdot \mathbb{P}_{S \sim \mathcal{D}}(S \notin \Omega)
\end{equation*}
Minimizing \( \ell(\mathcal{D}; G) \) encourages both low cost and feasibility, and guarantees that with high probability: 
a sample \( S_\ast \sim \mathcal{D} \) exists such that \( S_\ast \in \Omega \) and 
\( f(S_\ast; G) \leq \ell(\mathcal{D}; G)/(1 - t) \). 
\subsubsection{Integer Programming Formulations for ESPP}
Leveraging the PM for ESPP requires analytical characterizing the underlying optimization problem for ESPP. We begin by presenting the mixed-integer linear programming (MILP) formulation of ESPP~\cite{taccari2016integer} and highlighting its computational challenges. Let \( G = (V, E, w) \) be a directed graph with edge costs $w_{uv} \in \mathbb{R}$, each node must be visited at most once. $N^{+}(u)$ and $N^{-}(u)$ denote the sets of downstream and upstream nodes of node $u$, respectively, and $A(S)$ denotes the set of edges with both endpoints in $S\subseteq V$.
\begin{equation}
\min \quad \sum_{uv\in E} w_{uv}\,x_{uv} 
\label{eq:obj}
\end{equation}
\begin{equation}
\text{s.t.}\quad  \sum_{v\in N^{+}(u)} x_{uv} \;-\; \sum_{v\in N^{-}(u)} x_{vu}
   \;=\;
   \begin{cases}
     1  & \text{if } u=s,\\
    -1  & \text{if } u=t,\\
     0  & \text{o.w.}
   \end{cases}
    \label{flow}
\end{equation}
\begin{equation}
 \sum_{v\in N^{+}(u)} x_{uv} \;\le\; 1  \quad\forall\, u\in V
\label{degree}
\end{equation}
\begin{equation}
\sum_{uv\in A(S)} x_{uv} \;\le\; |S|-1,
\forall\, S\subseteq V \setminus \{s,t\},\; |S|\ge 2 
\label{subtour}
\end{equation}
\begin{equation} 
x_{uv}\in\{0,1\}  \quad\forall\, uv\in E 
\end{equation}
In the equations, Eq.~\eqref{flow} enforces flow conservation, and Eq.~\eqref{degree} ensures that the
outgoing degree of each node is at most one. Moreover, Eq.~\eqref{subtour} guarantees elementary visitations. However, a key challenge is that the number of subtour elimination constraints Eq.\eqref{subtour} grows exponentially with $|V|$. Theoretically, this formulation includes $\mathcal{O}(m)$ variables and $\mathcal{O}(2^{n})$ constraints.

While previous probabilistic method offers a principled way to characterize the existence of solutions, directly modeling the ESPP and incorporating them into loss function remain intractable due to the exponential number of subtour elimination constraints. To overcome this bottleneck, we introduce a learning perspective that derives tractable representations via GNNs, enabling efficient algorithmic reasoning. It bypasses hard constraints and trains a GNN whose output distribution concentrates on near-optimal, feasible solutions with high probability.
\section{A Probabilistic Method for the ESPP}
Although solving ESPP via DP is not applicable, it remains feasible to directly learn the optimal value function and then recover the path from that function. Let \( G = (V, E, w) \) be a directed graph, where \( w : E \to \mathbb{R} \) assigns a cost to each directed edge. We denote \( \mathcal{G} \) a set finite collection of graphs. We therefore develop our PM for ESPP by training an oracle algorithm operator at iteration $k^*$ denoted as  $F^{(k^*)}_\theta(G)$ such that it computes the value function $d_\theta(v)$ over nodes $V$. Utilizing the value function, we can further compute the edge probability over the edge set $E$. Each edge probability $(u,v) \in E$ is denoted as $p_{uv}$, which serves as the support for sampling ESPP paths.
This gives rise to the following loss function:
\begin{equation}
\mathcal{L}(\theta) = \mathbb{E}_{G \sim \mathcal{G}} \left[ 
\underbrace{\lambda_1 \left\| F^{(k^*+1)}_\theta(G) - F^{(k^*)}_\theta(G) \right\|}_{\text{(Value operator alignment)}} 
+ 
\underbrace{\lambda_2 (\Phi_\theta(G))}_{\text{(NCCs penalty)}} 
\right]
\label{eq:main}
\end{equation}

By minimizing $\mathcal{L}(\theta)$, Eq.~\eqref{eq:main} offers two distinct self-reinforcing benefits: (1) the oracle value function can be used to eliminate negative cycles, and (2) when negative cycles are eliminated with high probability, we can further cast standard SPP algorithms as algorithmic inductive biases that further facilitate the discovery of optimal value functions. Such construction of loss function directly empowers the development of an unsupervised learning framework for solving ESPP, which will be discussed in detail in the next section.  

More importantly, based on the surrogate loss function, we can formally establish the performance guarantee relates to the approximation errors of the learned value function and the NCCs penalty. To formally show this, we introduce the following three definitions, establish the high probability guarantee for NCCs, and bound the optimality gap. Detailed proofs can be found in Appendix A$\sim$E. 

\begin{definition}\label{opt-gap}
    \textbf{(Optimality gap)} 
    For any graph \( G \), we denote by \( \pi^*(G) \) the optimal path from source to sink, and by \( \pi_\theta(G) \) the path we obtained. The cost of a path is defined as the sum of edge weights along it. The optimality gap of a decoded path is defined as
    \begin{equation*}
    \Delta(G, \theta) := w(\pi_\theta(G)) - w(\pi^*(G)) \geq 0   
    \end{equation*}
    where \( w(\pi) \) denotes the total edge cost of path \( \pi \). 
\end{definition}
To enforce consistency with optimality principles, we define the following value function oracle algorithm operators.

\begin{definition}
    \textbf{(Value function operators)} Let $k^*$ denote the iteration at which an oracle algorithm (e.g., Bellman-Ford) converges, and \( d_\theta : V \to \mathbb{R} \) be the learned value function, where \( d_\theta(v) \) estimates the cost from node \( v \) to sink $t$. We define the operator $F_\theta^{(k^*)}(G)$ such that it yields the value function returned by the oracle algorithm after $k^*$ iterations on graph $G$. We take two iterations of the oracle algorithm operator as an illustrative example:
\begin{equation*}
F_\theta^{(1)}(G) =\min_{u \to v} \{ w(u, v) + d_\theta(v) \}
\end{equation*}
\begin{equation*}
F_\theta^{(2)}(G) = \min_{u \to w \to v} \{ w(u, w) + w(w, v) + d_\theta(v) \}.
\end{equation*} 
\end{definition}
Next, we will introduce a surrogate that approximates the presence of negative-cost cycles.
\begin{definition}\label{def:3}
    \textbf{(Negative-cost cycle surrogate)} Let value estimates $d_\theta(v) \in \mathbb{R}$ for each vertex $v \in V$ (w.r.t. a fixed sink) satisfying $d_\theta(u) \leq w_{uv} - d_\theta(v)$. The negative-cost cycle surrogate $\Phi_\theta(G)$ and slack $\delta_{uv}$ is further defined as:
\begin{equation*}
\Phi_\theta(G)=\;\frac{1}{\lvert E\rvert}\,\sum_{(u,v)\in E}\delta_{uv}
\;+\;\frac{1}{\lvert E\rvert}\,\sum_{(u,v)\in E} p_{uv}\,\delta_{uv}
\end{equation*} 
\begin{equation*}
\delta_{uv} = [\,d_\theta(u) - (w_{uv}+d_\theta(v))\,]_+ = \max(0,\; d_\theta(u) - w_{uv} - d_\theta(v))
\end{equation*}
\end{definition}

\begin{lemma}\label{lemma:1}
    \textnormal{\textbf{(Existence of positive slack in negative-cost cycles)}} If there exists a negative-cost cycle, then there must exist at least one edge in the cycle such that $\delta_{uv} > 0$.
\end{lemma}

Building on Lemma \ref{lemma:1}, we next establish Proposition \ref{pro:1}: a high-probability guarantee for non-negative-cost cycles.

\begin{proposition}\label{pro:1}
    \textnormal{\textbf{(The high probability of non-negative-cost cycles guarantee)}} Let \( G = (V, E, w) \) be a directed graph with edge costs $w_{uv} \in \mathbb{R}$, independent edge probabilities $p_{uv} \in [0, 1]$ for each edge $(u, v) \in E$, and value estimates $d_\theta(v) \in \mathbb{R}$ for each vertex $v \in V$ (w.r.t. a fixed sink) satisfying $d_\theta(u) \leq w_{uv} - d_\theta(v)$ .
    Let $\tilde{Z}$ be the random number of negative-cost cycles and $\mathcal{C}_{uv}^-$ denote the set of all negative-cost cycles that contain edge $(u,v)$, and define $M =\max_{(u,v)\in E} \sum_{C \in \mathcal{C}_{uv}^-} |C|$.
    Then for any $\varepsilon \in (0,1)$, if the penalty loss satisfies
    \begin{equation*}
    \Phi_{\theta}(G)\le\;\varepsilon_1
    \end{equation*}
    where $\varepsilon_1=\frac{\varepsilon}{M \lvert E\rvert}$, it holds that
    \begin{equation*}
    \Pr \left( \tilde{Z} = 0 \right) \geq 1 - \varepsilon.
    \end{equation*}
\end{proposition}
With this high-probability certificate, we further prove the existence of elementary shortest path under the condition if $G$ contains no negative-cost cycles.

\begin{proposition}
    \textnormal{\textbf{(Existence of ESPP)}} Let \( G = (V, E, w) \) be a directed graph with edge costs $w_{uv} \in \mathbb{R}$. If $G$ contains no negative-cost cycles, then for any source-sink pair $(s, t)$ connected by at least one path, there exists an elementary shortest path from $s$ to $t$.
    \label{pro:2}
\end{proposition}


\begin{lemma}
    \textnormal{\textbf{(Deterministic bound on optimality gap)}} There exist constants \( K_1, K_2 > 0 \); hyperparameters $\lambda_1$, $\lambda_2$ $>$ 0, such that for any graph \( G \) and parameters \( \theta \), if
    $\Phi_\theta(G) \le \varepsilon_1$ and $\left\| F^{(k^*+1)}_\theta(G) - F^{(k^*)}_\theta(G) \right\|\le \varepsilon_2$
    then the optimality gap of decoded path is bounded by:
    \begin{equation*}
    \Delta(G, \theta) \le K_1 \varepsilon_1 + K_2 \varepsilon_2.
    \end{equation*}
    Consequently, for \( K = \max\left( \frac{K_1}{\lambda_2}, \frac{K_2}{\lambda_1} \right) \), we have:
    \begin{equation*}
    \Delta(G, \theta)/K \le  L(G, \theta),
    \end{equation*}
    where 
    $L(G, \theta) = \lambda_1 \left\| F^{(k^*+1)}_\theta(G) - F^{(k^*)}_\theta(G) \right\| + \lambda_2 \Phi_\theta(G)$ represents the loss on $G$.
\end{lemma}


\begin{theorem}\label{the:1}
    \textnormal{\textbf{(The high-probability guarantee of near-optimality)}} The expected loss function is given by:
    \begin{equation*}
    \mathcal{L}(\theta) = \mathbb{E}_{G \sim \mathcal{G}} \left[
    \lambda_1 \left\| F^{(k^*+1)}_{\theta}(G) - F^{(k^*)}_{\theta}(G) \right\| + 
    \lambda_2 \Phi_{\theta}(G)
    \right]
    \end{equation*}
    Let \( \alpha, \beta, \gamma > 0 \) with $\alpha \in (0,1)$, $\beta \in (0,1)$. Suppose we train on \( n \) i.i.d. graphs \( G_1, \ldots, G_n \) to minimize empirical loss 
    \begin{equation*}
        \widehat{\mathcal{L}}_n(\theta) = \frac{1}{n} \sum_{i=1}^n L(G_i, \theta)
    \end{equation*}
    Let $\hat{\theta} = \arg\min_{\theta} \widehat{\mathcal{L}}_{n}(\theta)$ with \( \widehat{\mathcal{L}}_n(\hat{\theta}) \leq \epsilon \),
    If $n \geq R \cdot \frac{K^2 \log(1/\beta)}{\alpha^2 \gamma^2}$
    for a constant \( R \), then with probability \( \geq 1 - \beta \) over training data:
    \begin{equation*}
    \Pr \left( \Delta(G, \hat{\theta}) > \gamma \right) \leq \alpha.
    \end{equation*}
\end{theorem}


\begin{figure*}[t]
    \centering
    \includegraphics[height=0.30\linewidth]{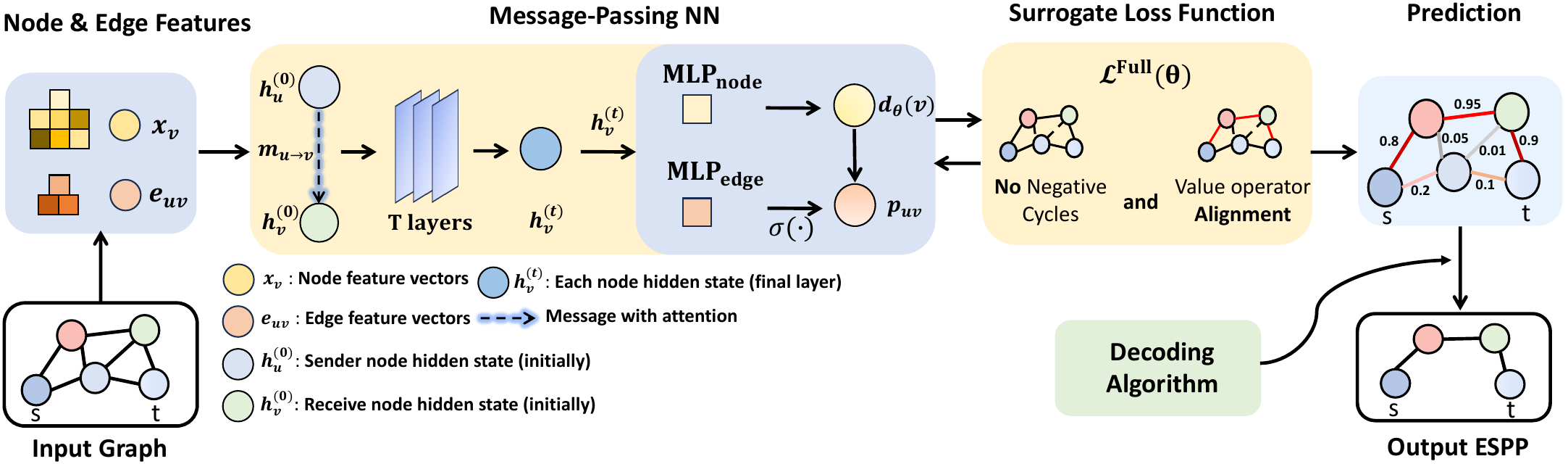}
    \caption{Unsupervised Learning for PM-based ESPP.}
    \label{fig:overview}
\end{figure*}


\section{Unsupervised Learning for PM-based ESPP}
Building on the PM-based formulation of ESPP, we next discuss how its theoretical performance guarantees can be effectively transformed into the development of an effective unsupervised learning framework. This framework comprises three key components, as illustrated in Figure~\ref{fig:overview}. First, we leverage the expressive power of GNN to build a baseline model that predicts value estimates. Second, we map the learned value function to a probabilistic distribution over edges, thus computing edge probabilities. The first two components in combination enables the training of the value function under the unsupervised learning framework. Finally, we introduce structural and algorithmic inductive biases into the loss function that aims to fully internalizes the self-reinforcing advantages of the proposed PM framework. Each component is detailed below.

\subsection{Baseline Model}
\subsubsection{Value Prediction via Message Passing}
We leverage a multi-hop message-passing framework that progressively transforms initial embedding $\mathbf{h}_v^{(0)}$ into value function estimations at the node level. For each message passing layer $t$, directional messages from node $u$ to node $v$ are computed via an edge-level multi-layer perceptron (MLP):
\begin{equation*}
\mathbf{e}_{u \rightarrow v}^{(t)} = \text{MLP}(\mathbf{h}_u^{(t)} \Vert \mathbf{e}_{u \rightarrow v}^{(t-1)} \Vert \mathbf{h}_v^{(t)})
\end{equation*}
Nodes then aggregate incoming messages from neighbors $u \in N(v)$ through an attention weight using edge features and message differences: 
\begin{equation*}
\mathbf{m}_{u \rightarrow v}^{(t)} = \sum_{u \in N(v)} \alpha_{u \rightarrow v}^{(t)} \cdot \text{MLP}(\mathbf{h}_u^{(t)} - \mathbf{h}_v^{(t)} \Vert \mathbf{e}_{u \rightarrow v}^{(t)})
\end{equation*}
The node embeddings are then updated using a minimum operation between the aggregated messages and previous embeddings, which structurally mirrors the Bellman operator for ESPP as:
\begin{equation*}
\mathbf{h}_v^{(t)} = \min(\mathbf{m}_{u \rightarrow v}^{(t)}, \mathbf{h}_v^{(t-1)})
\end{equation*}
After $T$ message-passing layers, the final node embedding of each node $\mathbf{h}_v^{(t)}$ are mapped to node value estimates $d_\theta(v)$ through a node MLP layer:
\begin{equation}
d_\theta(v) = \text{MLP}_{\text{node}}(\mathbf{h}_v^{(t)})
\label{eq:nodev}
\end{equation}

\subsubsection{Learning Value Prediction via Edge Probability}
The learned $d_\theta(v)$ in Eq.~\ref{eq:nodev}, by construction, does not represent physically meaningful value function estimations without additional training signals. This limitation can be effectively resolved by explicitly relating $d_\theta(v)$ to the ESPP formulation in Eq.~\ref{eq:obj} through edge probability $p_{uv}$. Specifically, we can transform node level $d_\theta(v)$ to edge level logits as $d_\theta(u)-d_\theta(v)$, that further maps to edge-level probability $p_{uv}$ using a sigmoid function:
\begin{equation*}
p_{uv} = \sigma(d_\theta(v) - d_\theta(u))
\end{equation*}
This mapping allows the definition of a  well-structured loss function within our unsupervised learning framework:
\begin{equation}
  \mathcal{L}_{\text{BASE}} = \sum_{uv} w_{uv}p_{uv} + \lambda_1 \mathcal{L}_{Flow}+\lambda_2 \Phi_\theta(G)
  \label{eq:vf}
\end{equation}
where the flow conservation constraints can be formalized as a violation penalty $\mathcal{L}_{\mathrm{flow}}$ over all the nodes as:
\begin{equation*}
\mathcal{L}_{\mathrm{Flow}}=\sum_{uv}p_{uv}-\sum_{vu}p_{vu}
=\begin{cases}
1 & \text{if } u=s,\\
-1 & \text{if } u=t,\\
0 & \text{o.w.}
\end{cases}
\end{equation*}
Note that the last term in Eq.~\eqref{eq:vf} is the same as the NCCs penalty as in Eq.~\eqref{eq:main} and the first two terms implicitly integrate the Bellman operator in Eq.~\eqref{eq:vf} with the mathematical models of ESPP by using $p_{uv}$ as a bridge. Based on this formulation, the resulting unsupervised learning model can be readily optimized via stochastic gradient descent. Upon convergence, the learned values $d_\theta(v)$ naturally align with meaningful value function estimations at each node. This baseline model further guarantees the effective elimination of NCCs with value function estimations, thus contributing to the optimality of the learned solution for the ESPP, as demonstrated in the previous section. 


\subsection{Incorporating Structural and Algorithmic Inductive Biases}
We next introduce important functional components that can be embedded into the baseline model that will significantly improve its performances. 
\subsubsection{Distributional alignment} In the baseline model, edge probabilities  $p_{uv}$ are learned independently for each edge $uv$. However, the probability of choosing an outgoing edge from a given node $u$ should be proportional to the downstream nodes' value function estimates. To explicitly enforce this structural alignment, we introduce a target transition probabilities $q_{uv}$ , computed as a differentiable softmax operator over the combined edge cost and the downstream node value estimate as:
\begin{equation*}
q_{uv} = \frac{\exp\left(-\frac{w_{uv} + d_\theta(v)}{\tau}\right)}{\sum_{(v' \in N^+(u)} \exp\left(-\frac{w_{uv'} + d_\theta({v'})}{\tau}\right)}
\end{equation*}
where $\tau$ is a temperature parameter that controls the softness of the target distribution. Note that the scales of distributions $p$ and $q$ may differ significantly, as $\sum_{v\in N^+(u)}p_{uv}$ could be considerably less than 1. To overcome this scale discrepancy, we employ a cosine similarity measure as the scale-invariant metric that effectively aligns the directional aspects of the two distributions. We achieve this by minimizing the following alignment loss:
\begin{equation*}
\mathcal{L}_{\text{DA}} = \frac{1}{|V|} \sum_{u \in V} \left[ 1 - \frac{ \sum_{v\in N^+(u)} p_{uv} q_{uv} }{ \sqrt{ \sum_{v\in N^+(u)} p_{uv}^2 } \cdot \sqrt{ \sum_{v\in N^+(u)} q_{uv}^2 } } \right]
\end{equation*}


\subsubsection{Dynamic Programming Alignment} The estimates of value function can be further tightened by explictly incorporating the DP operation over the specific graph topology. Specifically, the estimated value function should satisfy: 
\begin{equation*}
d_\theta(v) = \min_{u \in N^-(v)} \left\{ w_{uv} + d_\theta(u) \right\}
\end{equation*}
While the above Bellman operator is non-differentiable, we leverage the softmin operator 
\begin{equation*}
m_v = -\tau \log \sum_{u\in N^-(v)} \exp\!\left(-\frac{w_{uv}+d_\theta(u)}{\tau}\right)
\end{equation*}
where $\tau>0$ is a temperature parameter. This allows us to enforce the consistency between the learned value function estimates and the value induced from the estimated values of the neighbors, where we seek to minimize the consistency gap as:
\begin{equation*}
\mathcal{L}_{\mathrm{DPA}}=\frac{1}{|V|} \sum_{v \in V} (d_\theta(v) - m_v)^2 
\end{equation*}
\subsubsection{Algorithmic biases} With value function estimation and NCCs elimination established, we further enhance model generalization by explicitly embedding the iterative dynamics of the Bellman-Ford (BF) algorithm \cite{ford1956network,bellman1958routing}. Specifically, we explicitly unroll 
$T$ iterations of the BF algorithm as differentiable layers applied to the learned node values, using the same softmin relaxation as in the DP alignment:
\begin{equation*}
d_\theta(v)^{(T)} = -\frac{1}{\tau} \log \sum_{u\in N^-(v)}  \exp\left(-\tau(d_\theta(u)^{(T-1)} + w_{uv})\right)
\end{equation*}
with $d_\theta(v)^{(0)}=d_\theta(v)$. At convergence, the BF algorithm guarantees that the node values satisfy the Bellman optimality condition (i.e., they remain stable through successive iterations). Thus, when the estimated node values $d_\theta (v)$ remain unchanged after $T$ differentiable BF iterations, the optimality condition is effectively enforced. We capture this optimality condition explicitly by minimizing the following squared-error loss term:
\begin{equation*}
\mathcal{L}_{\text{AB}} = \frac{1}{|V|} \sum_{v \in V} \left(d_\theta(v)^{(T)} - d_\theta(v)\right)^2
\end{equation*}
This explicitly integrates a rigorous algorithmic prior into the training objective that simultaneous improve solution quality and enhance model generalization across diverse graph instances.

\subsubsection{Advantage baseline}
Our final modeling component explicitly addresses the high variability in solution structure across ESPP instances. Unlike the TSP, ESPP solutions may range from as few as a single edge to potentially involving every edge in the graph. Inspired by the role of advantage function in reinforcement learning~\cite{wang2016dueling}, we introduce an advantage baseline to stabilize the training process by reducing objective variance associated with this structural variability. Specifically, we construct the baseline by solving a relaxed ESPP without the subtour elimination constraints using \textsc{Gurobi}~\cite{gurobi} to obtain fractional solutions. These fractional solutions are then decoded into a heuristic elementary path of length $c^{\text{LP}}$. Formally, our advantage baseline loss component is:
\begin{equation*}
\mathcal{L}_{\mathrm{ADV}}=\sum_{uv\in E_G} w_{uv} p_{uv}- c^{\text{LP}}
\end{equation*}
This formulation explicitly encourages the learned solutions to outperform the heuristic baseline, improving training stability and facilitating better generalization across diverse ESPP scenarios.

\subsection{Integrated Surrogate Loss Function}
In summary, our surrogate loss for ESPP integrates two synergistic terms derived from Theorem~\ref{the:1}, with additional flow conservation to enforce path feasibility and an advantage baseline for variance reduction. The final integrated surrogate loss function is summarized below:

\begin{align*}
\mathcal{L}^{\text{Full}}(\theta) = \mathbb{E}_{G \sim \mathcal{G}} \Big[ & 
\underbrace{\mathcal{L}_{\mathrm{ADV}} + 
\lambda_1\mathcal{L}_{\mathrm{Flow}}}_{\text{Value operator alignment}}+ \underbrace{\lambda_2 \Phi_\theta(G)}_{\text{NCCs penalty}} \\
& + \underbrace{\lambda_3(\mathcal{L}_{\mathrm{DA}} + \mathcal{L}_{\mathrm{DPA}} + \mathcal{L}_{\text{AB}})}_{\text{Inductive biases}}
\Big]
\end{align*}
Theoretically, sufficiently large penalty weights $\lambda_1$, $\lambda_2$, and $\lambda_3$ can enforce the corresponding constraints via soft penalty minimization. In practice, one would benefit from tuning them as hyper-parameters to balance constraint satisfaction and optimization stability.

\subsection{Decoding Algorithm}
Once the GNN has produced a probability \(p_{uv}\), we recover an elementary \(s\to t\) path by a polynomial-time sequential decoding algorithm from the learned distribution. For each of \(N\) trials, we initialize a path $P$ and a visited set $V_{\mathrm{visited}}$. At the current node \(u\), we form the eligible outgoing set
$O_u =\bigl\{(u,v)\in E : v\notin V_{\mathrm{visited}}\bigr\}$.
In this way, the decoder get edge probabilities from the \textsc{Espp-nnaa} while maintaining the ability to guarantee an elementary, low‐cost path via sampling. The detailed decoding algorithm is shown in \textbf{Algorithm 1} in Appendix F.






\section{Experiments}
The experiments are primarily conducted on Erdős--Rényi (ER) graphs~\cite{erdds1959random}, which are instantiated at node sizes $N\in\{30,50,100\}$ and initialized with random edge weights $\mathcal{U}_{[-1,1]}$, including various configurations with NCCs. To evaluate generalization, we further test the trained model on two different graph families: Grid and Barabási--Albert (BA) graphs~\cite{albert2002statistical}. The synthetic datasets provide controlled topological diversity. They span random connectivity, spatial regularity, and scale free structures to rigorously test generalization beyond training distributions. Each dataset includes 2,000 instances and is split into training (70\%), test (20\%), and validation (10\%) sets. All the learning-based methods are trained via Adam optimizer ($\text{lr}=10^{-3}$) on mini‑batches of 64 graphs. 
\subsection{Model Comparison} 
The performance of our proposed \textsc{Espp-nnaa} is evaluated by comparing it against unsupervised baselines, exact solvers, and classic heuristics.

\textbf{\textsc{Espp-aa}}\& \textbf{\textsc{Espp-nn}:} We evaluate two variants, with \textsc{Espp-aa} focusing exclusively on algorithm alignment and \textsc{Espp-nn} emphasizing negative-cost cycle reduction. 

\textbf{\textsc{Espp-erdős}} \& \textbf{{\textsc{Espp-heatmap}}:} \textsc{Espp-erdős} adopts the traditional probabilistic approach \cite{karalias2020erdos} that models the ESPP solely via a surrogate loss. Specifically, we employ a relaxed formulation proposed in the integer programming \cite{taccari2016integer}. Additionally, \textsc{Espp-heatmap} is the baseline that we further utilize the adjusted-heatmap surrogate from the formulation in \cite{abouelrous2025graph}.

\textbf{\textsc{Spp-erdős}:} This is the \textsc{Espp-erdős} variant without elementary constraints. 

\textbf{\textsc{Labeling algorithm}:} We use the \textsc{labeling algorithm} \cite{feillet2004exact} for gaining exact solution of ESPP.

\textbf{\textsc{Beam Search}:} The beam search heuristic \cite{lowerre1976harpy} limits the number of partial paths explored at each expansion step. We use \textsc{Beam Search} as the baseline. 

\textbf{\textsc{LP-Heuristic}}\&\textbf{\textsc{Randomized-algorithm}:} \textsc{LP-Heuristic} solves MILP model of ESPP without subtour elimination constraints using \textsc{Gurobi}~\cite{gurobi}. Lastly, we include a \textsc{randomized algorithm}, using the same decoding algorithm but with equal probability over outgoing edges. This helps quantify the quality of our \textsc{Espp-nnaa} output edge probabilities.
\begin{table}[t]
\centering
\small
\setlength{\tabcolsep}{1mm}
{
\begin{tabular}{@{}lcc|cc|cc@{}}
\toprule
\textbf{Algorithm} & \multicolumn{2}{c|}{\textbf{30-node}} & \multicolumn{2}{c|}{\textbf{50-node}} & \multicolumn{2}{c}{\textbf{100-node}} \\
 & \% gap & t [s] & \% gap & t [s] & \% gap & t [s] \\
\midrule
\textsc{\textbf{Espp-nnaa}} & \textbf{-20.42} & \textbf{1.88} & \textbf{-78.48} & \textbf{3.02} & \textbf{-222.96} & \textbf{6.31} \\
\textsc{Espp-aa} & -17.55 &1.93& -75.19
&2.85&-219.69&6.20\\
\textsc{Espp-nn} & 3.92 &0.93 &-28.58&1.32&-71.56&1.90\\
\textsc{ESPP-Erdős} & 2.43 & 0.87 &-22.07&1.17&-71.20&1.90\\
\textsc{ESPP-Heatmap} & 37.33 &1.33 &23.77&1.72&7.08&2.44\\
\textsc{SPP-Erdős} &3.87&0.86&-19.42&1.13&-67.63&1.90\\
LP-Heuristic & 11.30 & 0.58 & -39.89 & 1.30 & -166.12 & 5.40 \\
Randomized & 57.63 & 0.62 & 55.27 & 0.76 & 53.33 & 1.13 \\
\midrule
Labeling (Exact) & -25.54 & 7.04 & -95.91 & 310.92 & / & $>$7200 \\
Beam (Baseline) & 0.00 & 5.04 & 0.00 & 9.62 & 0.00 & 23.40 \\
\bottomrule
\end{tabular}
}
\caption{Comparison of methods on graph test set}
\label{tab:model_comparison}
\end{table}
We summarize results on ER graphs in Table~\ref{tab:model_comparison}. Among the unsupervised baselines, \textsc{Espp-nnaa} achieves the strongest gap reduction across graph sizes by synergistically combining algorithm alignment with feasibility control. \textsc{Espp-aa} remains competitive, likely due to its value-function shaping toward near-optimal routes and enforced elementarity during decoding. Without this guidance, \textsc{Espp-nn}, \textsc{Espp-erdős}, and \textsc{Espp-heatmap} stall in local optima. Ablating elementarity \textsc{Spp-erdős} degrades performance. From the standpoint of exact solvers and classical heuristics, traditional \textsc{label algorithm} quickly deteriorates as the graph grows. The running time exceeds one day for $100$-node graphs. \textsc{Beam search} alleviates this time issue, yet it trades path quality for speed.
\textsc{LP-Heuristic} yields large gaps on small graphs; plateaus deterministically and cannot push the gap any lower.

\begin{figure}[htbp]
    \centering
    \begin{subfigure}[t]{0.41\linewidth}
        \centering
        \includegraphics[width=\linewidth]{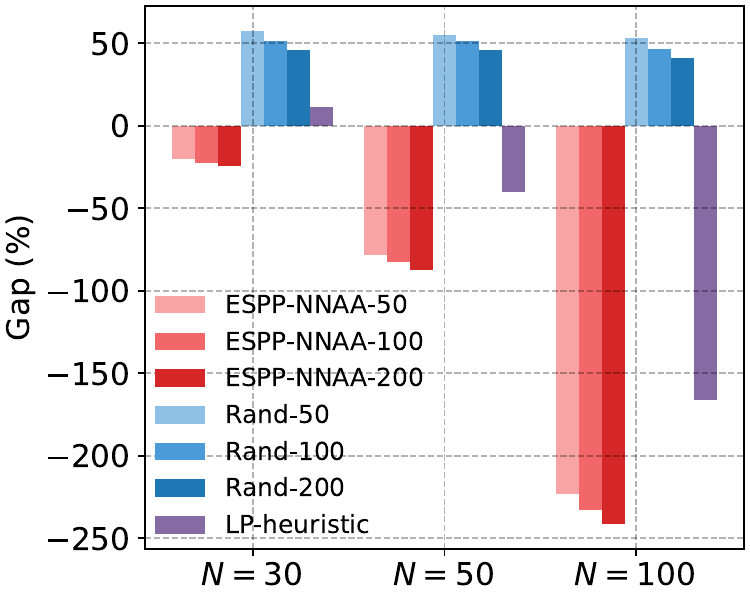}
        \caption{Sampling Study }
        \label{fig:sampling_study}
    \end{subfigure}
    \hfill
    \begin{subfigure}[t]{0.58\linewidth}
        \centering
        \includegraphics[width=\linewidth]{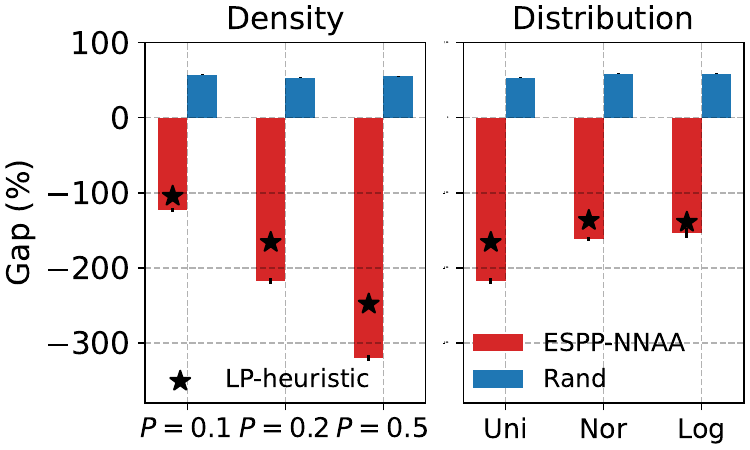}
        \caption{Density \& Distribution Study}
        \label{fig:density_distribution}
    \end{subfigure}
    \caption{Studies on Sampling, Density, and Distribution.}
    \label{fig:combined_studies}
\end{figure}


Importantly, our sampling study (Figure \ref{fig:sampling_study}) shows that increasing decoding samples from 50 to 200 consistently reduces optimality gaps. At 200 samples, gaps closely match exact solutions for $|V|$=30/50 with polynomial runtime growth. 50-100 samples offer the most time and quality trade-off. In extended experiments on 100-node ER graphs (Figure \ref{fig:density_distribution}), \textsc{Espp-nnaa} consistently outperformed all baselines across varied densities and edge-weight distributions(Uniform; Normal; Lognormal). Under sparse to moderately dense topologies (\( p = 0.1, 0.2, 0.5 \)), it maintained great negative gaps, exceeding heuristics substantially even in the most challenging dense graph case ($p=0.5$).

\begin{table}[htbp]
\centering
\small
\begin{tabular}{@{}llcc@{}}
\toprule
$|V|$ & Graph type & \textsc{LP-Heuristic} gap & \textsc{Espp-nnaa} gap \\
\midrule
\multirow {3}{*}{30}
  & ER   &  11.30±0.00   & \textbf{-16.96±0.47} \\
  & Grid & 35.22±0.00  & \textbf{-8.14±15.96} \\
  & BA   &  7.02±0.00 & \textbf{-21.94±0.44} \\[0.3em]

\multirow {3}{*}{50}
  & ER   & -39.89±0.00  & \textbf{-76.53±0.84} \\
  & Grid & 5.53±0.00  & \textbf{-54.42±13.76} \\
  & BA   & -41.79±0.00  & \textbf{-82.71±1.50} \\[0.3em]

\multirow {3}{*}{100}
  & ER   & -166.12±0.00  & \textbf{-185.03±4.18} \\
  & Grid & -59.62±0.00 & \textbf{-122.65±9.11} \\
  & BA   & -179.83±0.00 & \textbf{-185.45±4.65} \\
\bottomrule
\end{tabular}
\caption{Generalization ability: gap (\%) vs. beam search}
\label{tab:generation-er-only}
\end{table}


\subsection*{Generalization Performance}
Table \ref{tab:generation-er-only} demonstrates \textsc{Espp-nnaa}'s cross-topology generalization capability. Trained exclusively on ER graphs at node sizes $|V|\in\{30,40,50\}$, it maintains high performance on unseen BA and Grid structures across sizes. The model shows particular strength on BA networks. For Grid topologies, geometric structure cause initial variance, but performance improves with graph size grows. This structural adaptability confirms algorithmic routing capabilities beyond its training topologies, exemplifying neural algorithmic reasoning.

\begin{figure}[htbp]
  \centering
  \begin{subfigure}[b]{0.48\linewidth}
    \includegraphics[width=\linewidth]{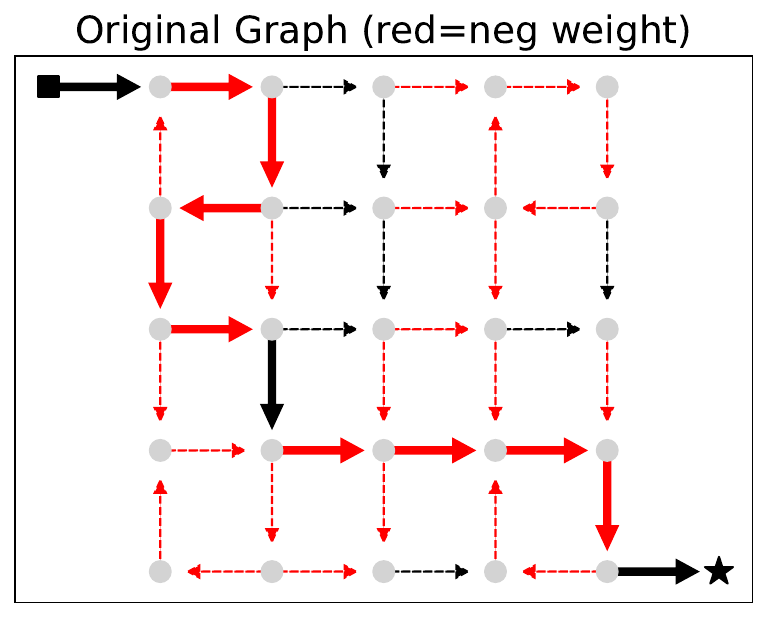}
    \caption{Original Graph}
    \label{fig:demo:raw}
  \end{subfigure}
  \hfill
  \begin{subfigure}[b]{0.48\linewidth}
    \includegraphics[width=\linewidth]{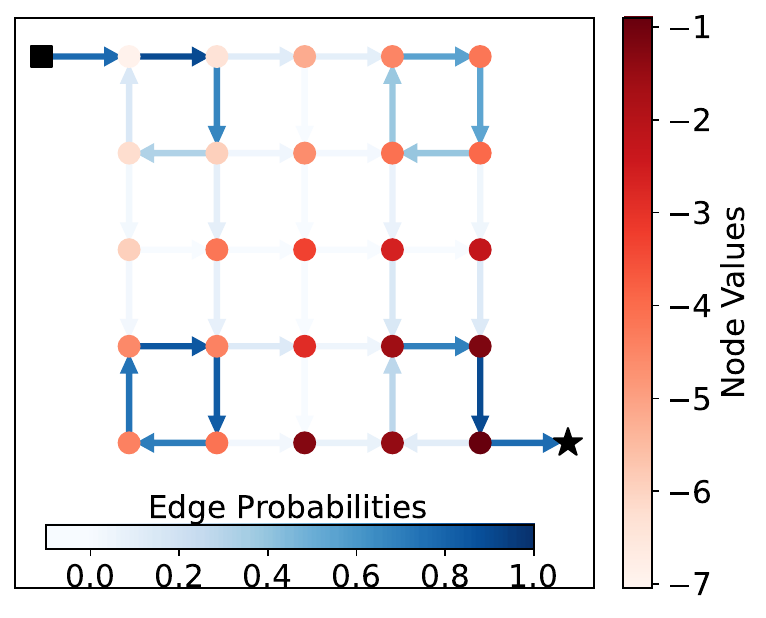}
    \caption{Low Penalty}
    \label{fig:demo:neg1}
  \end{subfigure}

  \begin{subfigure}[b]{0.48\linewidth}
    \includegraphics[width=\linewidth]{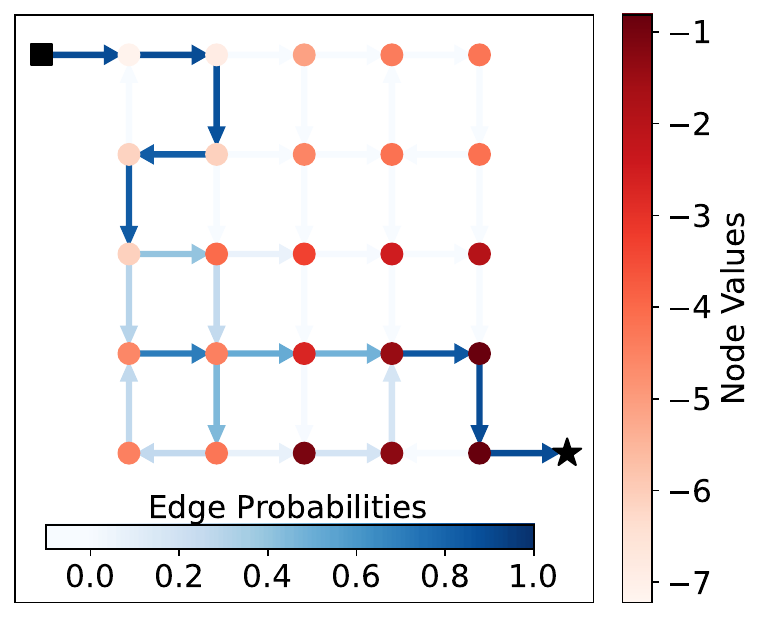}
    \caption{Medium Penalty}
    \label{fig:demo:neg2}
  \end{subfigure}
  \hfill
  \begin{subfigure}[b]{0.48\linewidth}
    \includegraphics[width=\linewidth]{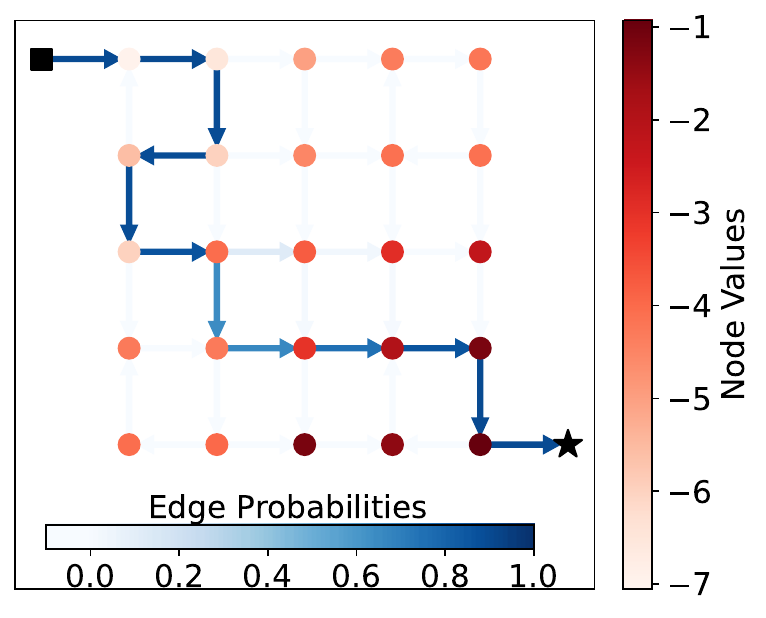}
    \caption{High Penalty}
    \label{fig:demo:neg3}
  \end{subfigure}

  \caption{(a) Original graph with negative weights (in red). (b–d) Learned edge probabilities and node value estimates as we progressively increase the penalties (NCCs penalty, Flow conservation penalty).}
  \label{fig:demo}
\end{figure}

\subsection{Performance Visualization}

Using grid graphs (Figure \ref{fig:demo:raw}-\ref{fig:demo:neg3}), we demonstrate how NCCs and flow conservation penalties collectively control NCCs elimination, enforce flow constraints, and regulate bias–variance trade‑offs. The original graph (Figure \ref{fig:demo:raw}) shows negative edges (red) and ground-truth paths. Under low penalties (Figure \ref{fig:demo:neg1}), 
the model primarily relying on algorithm alignment, often gets trapped in negative-cost cycles. Medium penalties (Figure \ref{fig:demo:neg2}) achieve equilibrium: edge probabilities form acyclic ESPP subgraphs while node values develop monotonic sink gradients. However, under high penalties (Figure \ref{fig:demo:neg3}), the graph becomes overly deterministic, exhibiting high bias and low variance, which hinders sampling-based decoding. Thus, a medium penalty provides the optimal balance for our decoding algorithm.

\begin{table}[]
\small
\centering
\begin{tabular}{lcc}
\toprule
Component & \textbf{Gap (\%)}$\downarrow$ & \textbf{Path Length}$\downarrow$ \\
\midrule
\multicolumn{3}{c}{\textbf{$|V| = 50$}}                                          \\ \midrule
Full model              & \textbf{-76.67$\pm$1.06}  & \textbf{-10.72$\pm$0.06} \\
w/o DPA              & -75.78$\pm$3.27           & -10.67$\pm$0.20          \\
w/o DA          & -3.34$\pm$2.22            & -6.27$\pm$0.14           \\
w/o Advantage            & -36.96$\pm$3.44           & -8.31$\pm$0.21           \\
w/o AB & -70.46$\pm$7.35           & -10.34$\pm$0.44          \\ \midrule
\multicolumn{3}{c}{\textbf{$|V| = 100$}}                                         \\ \midrule
Full model              & \textbf{-217.53$\pm$3.52} & \textbf{-19.15$\pm$0.21} \\
w/o DPA              & -164.40$\pm$49.83         & -15.95$\pm$3.00          \\
w/o DA           & -6.84$\pm$51.59           & -6.44$\pm$3.11           \\
w/o Advantage            & -49.63$\pm$2.47           & -9.02$\pm$0.15           \\
w/o AB & -182.64$\pm$40.56         & -17.04$\pm$2.44          \\ 
\bottomrule
\end{tabular}
\caption{Ablation study on \textsc{Espp-nnaa}}
\label{tab:ablation}
\end{table}



\subsection{Ablation Study}

Table~\ref{tab:ablation} shows leave-one-out ablations for \( |V| \)=50 and \( |V| \)=100. The full model achieves the largest negative gap, the elementary shortest path length, and the smallest standard deviation. Ablating the distributional alignment loss causes the most severe degradation: the gap deteriorates from  -76\%  to -3\%  (\( |V| \)=50) and -217\% to -6\%  (\(|V|\)= 100), confirming its critical role in maintaining directional consistency with Bellman principle. Ablating the advantage baseline, dynamic programming alignment and algorithmic biases also substantially degrades performance, with impact magnifying at larger graph sizes.
All performance drops intensify at \( |V| = 100 \), highlighting growing component synergy with search space complexity.

\section{Conclusion}
This is the first work to employ unsupervised learning for solving the ESPP with negative-cost cycles. Our model simultaneously learns value estimates and edge probabilities via a certified surrogate loss function. The results demonstrate that the learned edge distributions effectively reduce the original graphs into non-negative cycle subgraphs, while simultaneously capturing the principles underlying shortest-path algorithms. Empirical evaluations confirm that our method is competitive with unsupervised baselines and traditional heuristics in both performance and efficiency, particularly as graph sizes grow. Future extensions may consider resource constraints and explore tighter alignment with classical solvers, and richer inductive biases.

\bibliography{main}
\end{document}